\documentclass[journal]{IEEEtran}

\usepackage{graphicx}
\usepackage[cmex10]{amsmath}
\usepackage{algorithmic}
\usepackage{algorithm}
\usepackage{array}
\usepackage[nocompress]{cite}
\usepackage[hyphens]{url}
\usepackage[table]{xcolor}
\usepackage{bbm}
\usepackage{epsfig}
\usepackage{enumitem}
\usepackage{bm}
\usepackage{amsfonts,amssymb,multirow,bigstrut,booktabs,ctable,latexsym}
\usepackage{subfigure}
\usepackage{mathtools}
\usepackage[mathscr]{eucal}
\usepackage{verbatim}
\usepackage{amsthm}
\usepackage{algorithm}
\usepackage{color}
\usepackage{xcolor}
\usepackage{physics}
\usepackage{soul}
\usepackage{subfloat}
 
\newtheorem{theorem}{Theorem}
\newtheorem{corollary}{Corollary}
\newtheorem{assumption}{Assumption}
\newtheorem{lemma}{Lemma}
\newtheorem{definition}{Definition}



\IEEEoverridecommandlockouts

\begin{document}
\title{Concentrated Differentially Private and Utility Preserving Federated Learning}
  \author{Rui~Hu,~\IEEEmembership{Student Member,~IEEE,} Yuanxiong~Guo,~\IEEEmembership{Senior~Member,~IEEE}
          and~Yanmin~Gong,~\IEEEmembership{Member,~IEEE}
 \thanks{R. Hu and Y. Gong are with the Department of Electrical and Computer Engineering, University of Texas at San Antonio, San Antonio,
 TX, 78249 USA (e-mail: \{rui.hu, yanmin.gong\}@utsa.edu).}
 \thanks{Y. Guo is with the Department of Information Systems and Cyber Security, University of Texas at San Antonio, San Antonio, TX, 78249 USA (e-mail: yuanxiong.guo@utsa.edu).}
 }
\maketitle

\begin{abstract}
Federated learning is a machine learning setting where a set of edge devices collaboratively train a model under the orchestration of a central server without sharing their local data. At each communication round of federated learning, edge devices perform multiple steps of stochastic gradient descent with their local data and then upload the computation results to the server for model update. During this process, the challenge of privacy leakage arises due to the information exchange between edge devices and the server when the server is not fully trusted. While some previous privacy-preserving mechanisms could readily be used for federated learning, they usually come at a high cost on convergence of the algorithm and utility of the learned model. %
In this paper, we develop a federated learning approach that addresses the privacy challenge without much degradation on model utility through a combination of local gradient perturbation, secure aggregation, and zero-concentrated differential privacy (zCDP). We provide a tight end-to-end privacy guarantee of our approach and analyze its theoretical convergence rates. Through extensive numerical experiments on real-world datasets, we demonstrate the effectiveness of our proposed method and show its superior trade-off between privacy and model utility.
\end{abstract}
\IEEEpeerreviewmaketitle

\section{Introduction}\label{sec:intro}
With the development of Internet-of-Things (IoT) technologies, smart devices with built-in sensors, Internet connectivity, and programmable computation capability have proliferated and generated huge volumes of data at the network edge over the past few years.
These data can be collected and analyzed to build machine learning models that enable a wide range of intelligent services, such as personal fitness tracking\cite{iotwearable}, traffic monitoring\cite{iottraffic}, smart home security\cite{smarthome}, and renewable energy integration\cite{iotenergy}. 
However, data are often sensitive in many services and can leak a lot of personal information about the users. Due to the privacy concern, users could be reluctant to share their data, prohibiting the deployment of these intelligent services. 

\textit{Federated Learning} is a novel machine learning paradigm where a group of edge devices collaboratively learn a shared model under the orchestration of a central server without sharing their local data. It mitigates many of the privacy risks resulting from the traditional, centralized machine learning paradigm, and has received significant attention recently \cite{kairouz2019advances}. At each communication round of federated learning, edge devices download the shared model from the server and compute updates to it using their own datasets, and then these updates will be gathered by the server to update the shared model. Although only model updates are transmitted between edge devices and the server instead of the raw data, such updates could contain hundreds of millions of parameters in modern machine learning models such as deep neural networks, resulting in high bandwidth usage per round. Moreover, many learning tasks require a large number of communication rounds to achieve a high model utility, and hence the communication of the whole training process is expensive. Since most edge devices are resource-constrained, the bandwidth between the server and edge devices is rather limited, especially {in} up-link transmissions. Therefore, in the state-of-the-art federated learning algorithms, each edge device would perform {multiple} local iterations in each round to obtain a more accurate model update, so that the total number of communication rounds to achieve a desired model utility will be reduced.

Besides communication overhead, federated learning faces several additional challenges, among which privacy leakage is a major one\cite{kairouz2019advances}. 
Although in federated learning edge devices keep their data locally and only exchange ephemeral model updates which contain less information than raw data, this is not sufficient to guarantee data privacy. For example, by observing the model updates from an edge device, it is possible for the adversary to recover the private dataset in that device using reconstruction attack \cite{al2016reconstruction} or infer whether a sample is in the dataset of that device using membership inference attack \cite{shokri2017membership}. Especially, if the server is not fully trusted, it can easily infer the private information of edge devices from the received model updates during the training by employing existing attack methods. Therefore, how to protect against those advanced privacy attacks and provide rigorous privacy guarantee for each device in federated learning without a fully trusted server is challenging and needs to be addressed. 


In order to motivate and retain edge devices in federated learning, it is desirable to achieve data privacy guarantee for devices. 
Several prior efforts have considered privacy \cite{bonawitz2016practical,geyer2017differentially,mcmahan2018learning,liang2020exploring,wei2020federated} in federated learning. \cite{bonawitz2016practical} proposed a secure aggregation method to protect model updates during the training of federated learning, which is orthogonal to our work. Differential privacy (DP) is adopted in \cite{geyer2017differentially,mcmahan2018learning,liang2020exploring} to protect the privacy of devices in federated learning, but they assumed a trusted server that adds noise into the aggregated model updates on behalf of the devices and did not provide any performance guarantee. Wei et al. \cite{wei2020federated} proposed a differentially private federated learning scheme with a trusted server and provided a performance guarantee only for convex models. Moreover, they used $(\epsilon,\delta)$-DP and its simple composition property to capture the total privacy loss across multiple iterations, which is loose and requires a large amount of noise to add, sacrificing mode utility.  


In this paper, we propose a novel federated learning scheme that protects the privacy of devices with high model utility, without assuming a fully trusted server. To provide a rigorous DP guarantee for each device without a fully trusted server, we leverage gradient perturbation that asks each device to add noise to perturb its gradient before uploading. However, when combining with communication-reduction strategy of federated learning directly, gradient perturbation adds too much noise to the model updates and leads to low model utility. To improve the model utility, we integrate a secure aggregation protocol with low communication overhead with the gradient perturbation to reduce the added noise magnitude. Furthermore, we utilize the zero-concentrated differential privacy (zCDP) to tightly capture the end-to-end privacy loss, so that less noise will be added under the same DP guarantee. 
Besides, our approach has a rigorous performance guarantee for both strongly-convex and non-convex loss functions.

In summary, the main contributions of this paper are as follows.
\begin{itemize}
    \item We propose a novel federated learning scheme for differentially private learning over distributed data without a fully trusted server. Our approach can rigorously protect the data privacy of each device with only marginal degradation of the model utility by integrating secure aggregation and gradient perturbation with Gaussian noise techniques. 
    \item We tightly compute the end-to-end privacy loss of our approach using zCDP, taking into account the privacy amplification effects of data subsampling, partial device participation and secure aggregation. Compared with using traditional $(\epsilon,\delta)$-DP and its simple composition property to count the privacy loss, our approach enables devices to add less noise and hence improves the model utility under the same privacy guarantee.
    
    \item We prove the convergence results of our approach for both strongly-convex and non-convex loss functions. Our approach achieves a convergence rate of $\mathcal{O}(1/\sqrt{n\tau T}) + \mathcal{O}({\tau \sigma^2}/{T})$ where $n$ is the number of devices, $T$ is the number of communication rounds, $\tau$ is the local iteration period, $\sigma^2$ is the variance of Gaussian noise added into the gradients at each local iteration. When $T$ is large and the first term dominates, our approach obtains the same asymptotic convergence rate as the classic non-private federated learning algorithm. 
    
    
    \item We conduct extensive evaluations based on the real-world dataset to verify the effectiveness of our approach. 
\end{itemize}

The rest of the paper is organized as follows. Preliminaries on privacy notations used in this paper are described in Section~\ref{sec:pre}. Section~\ref{sec:sys-mod} introduces the system setting and problem formulation, and Section~\ref{sec:DP-Fed} presents our private federated learning scheme. The privacy guarantee and convergence properties of our approach are rigorously analyzed in Section~\ref{sec:privacy_analysis} and Section~\ref{sec:convg}, respectively. Section~\ref{sec:exp} shows the evaluation results based on the real-world dataset. Finally, Section~\ref{sec:related} describes the related works, and Section~\ref{sec:con} concludes the paper.

\section{Preliminaries}\label{sec:pre}


In what follows, we briefly describe the basics of DP and their properties. 
DP is a rigorous notion of privacy and has become the de-facto standard for measuring privacy risk. 
%
$(\epsilon, \delta)$-DP \cite{dwork2014algorithmic} is the classic DP notion with the following definition:

\begin{definition}[$(\epsilon,\delta)$-DP]\label{DP} 
A randomized algorithm $\mathcal{M}$ is $(\epsilon,\delta)$-differentially private if for any two adjacent datasets $D, D^{\prime} \subseteq \mathcal{D}$ that have the same size but differ in at most one data sample and any subset of outputs $\mathcal{S} \subseteq range(\mathcal{M})$, it satisfies that:
\begin{equation}
\Pr[\mathcal{M}(D) \in \mathcal{S}] \leq e^{\epsilon} \Pr[\mathcal{M}(D^{\prime}) \in \mathcal{S}] + \delta.
\end{equation}
\end{definition}

The above definition reduces to $\epsilon$-DP when $\delta=0$. Here the parameter $\epsilon$ is also called the privacy budget. Given any function $f$ that maps a dataset $D\in\mathcal{D}$ into a scalar ${o}\in \mathbb{R}$, we can achieve $(\epsilon,\delta)$-DP by adding Gaussian noise $\mathcal{N}(0,\sigma^2)$ to the output scalar ${o}$, where the noise magnitude $\sigma$ is proportional to the sensitivity of $f$, given as $\Delta_2(f):= \|f(D) - f(D^\prime) \|_2$.

$\rho$-zCDP\cite{bun2016concentrated} is a relaxed version of $(\epsilon, \delta)$-DP. zCDP has a tight composition bound and is more suitable to analyze the end-to-end privacy loss of iterative algorithms. To define zCDP, we first define the privacy loss. Given any subset of outputs $ \mathcal{S} \in range(\mathcal{M}$), the privacy loss $Z$ of the mechanism $\mathcal{M}$ is a random variable defined as:
\begin{equation}
Z:= \log\frac{\Pr[\mathcal{M}(D) = \mathcal{S}]}{\Pr[\mathcal{M}(D^{\prime})=\mathcal{S}]}.
\end{equation}
zCDP imposes a bound on the moment generating function of the privacy loss $Z$. Formally, a randomized mechanism $\mathcal{M} $ satisfies $\rho$-zCDP if for any two adjacent datasets $D, D^{\prime} \subseteq \mathcal{D}$, it holds that for all $\alpha\in(1,\infty)$,
\begin{equation}\label{zcdp-bound}
\mathbb{E}[e^{(\alpha-1)Z}] \leq e^{(\alpha-1)\rho}.
\end{equation}
Here, \eqref{zcdp-bound} requires the privacy loss $Z$ to be concentrated around zero, and hence it is unlikely to distinguish $D$ from $D^{\prime}$ given their outputs. zCDP has the following properties \cite{bun2016concentrated, yu2019differentially}:
\begin{lemma}\label{rho-zcdp}
Let $f:\mathcal{D}\rightarrow\mathbb{R}$ be any real-valued function with sensitivity $\Delta_2(f) $, then the Gaussian mechanism, which returns $f(D) + \mathcal{N}(0,\sigma^2)$, satisfies $\Delta_2(f)^2/(2\sigma^2)$-zCDP.
\end{lemma}
\begin{lemma}\label{composition}
Suppose two mechanisms satisfy $\rho_1$-zCDP and $\rho_2$-zCDP, then their composition satisfies $\rho_1+\rho_2$-zCDP.
\end{lemma}
\begin{lemma}\label{zcdp-dp}
If $ \mathcal{M}$ is a mechanism that provides $\rho$-zCDP, then $ \mathcal{M}$ is $(\rho + 2\sqrt{\rho\log({1}/{\delta})}, \delta) $-DP for any $\delta > 0$.
\end{lemma}
\begin{lemma}\label{zcdp-shuffling}
Suppose that a mechanism $\mathcal{M}$ consists of a sequence of $E$ adaptive mechanisms, $\mathcal{M}_1,\dots,\mathcal{M}_E$, where each $\mathcal{M}_j$ satisfies $\rho_j$-zCDP ($1\leq j\leq E)$. {Let ${D}_1, \dots,{D}_E$ be the result of a randomized partitioning of the dataset ${D}$.} The mechanism $\mathcal{M}(D) = (\mathcal{M}_1( {D}_1),\dots,\mathcal{M}_E({D}_E))$ satisfies $\max_j \rho_j$-zCDP.
\end{lemma}


\section{System Modeling and Problem Formulation}\label{sec:sys-mod}

\subsection{Federated Learning System}
Consider a federated learning setting that consists of a central server and $n$ devices which are able to communicate with the server. Each device $i \in [n]$ has collected a local dataset $D_i=\{\xi_1^i, \dots, \xi_m^i\}$ of $m$ datapoints. The devices want to collaboratively learn a shared model $\bm{\theta}\in\mathbb{R}^d$ under the orchestration of the central server. 
Due to the privacy concern and high latency of uploading all local datapoints to the server, federated learning allows devices to train the model while keeping their data locally. Specifically, the shared model $\bm{\theta}$ is learned by minimizing {the overall empirical risk} on the union of all local datasets, that is,
\begin{equation}
\label{fed_obj}
    \min_{\bm{\theta}} f(\bm{\theta}) := \frac{1}{n}\sum_{i=1}^{n}  f_{i}(\bm{\theta}) \text{ with } f_{i}(\bm{\theta}) := \frac{1}{m}\sum_{\xi_i\in{D}_i} l(\bm{\theta}, \xi_i).
\end{equation}
Here, $f_i$ represents the local objective function of device $i$, $l(\bm{\theta};\xi_i)$ is the loss of the model $\bm{\theta} $ at a datapoint $\xi_i$ sampled from local dataset $D_i$.

\begin{figure}[t]
\centering
\includegraphics[width=3in]{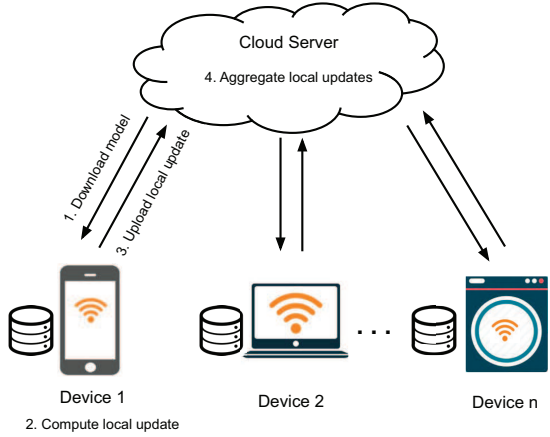}
\caption{System architecture of federated learning.}
\label{fig:system_model}
\vspace*{-10pt}
\end{figure}

In federated learning, the central server is responsible for coordinating the training process across all devices and maintaining the shared model $\bm{\theta}$. The system architecture of federated learning is shown in Fig.~\ref{fig:system_model}. At the beginning of each iteration, devices download the shared model $\bm{\theta}$ from the server and compute local updates on $\bm{\theta}$ using their local datasets. Then, each device uploads its computed result to the server, where the received local results are aggregated to update the shared model $\bm{\theta}$. This procedure repeats until certain convergence criteria are satisfied.

The classic approach to solve Problem~\eqref{fed_obj} is the federated averaging (FedAvg) algorithm\cite{mcmahan2017communication}. In FedAvg, the server first selects a subset of devices uniformly at random and then lets the selected devices perform multiple iterations of SGD to minimize the local objectives before sending their local computation results to the server. Let $\tau$ represent the local iteration period and $t\in[0,\dots, T-1]$ represent the index of communication round. Specifically, at round $t$, a set $\Omega_t$ of $r$ devices are selected to download the current shared model $\bm{\theta}^t$ from the server and perform $\tau$ local iterations on $\bm{\theta}^t$. Let $\bm{\theta}_i^{t,s}$ denote the local model of device $i$ at $s$-th local iteration of the $t$-th round. At each local iteration $s=0,\dots,\tau-1$, device $i$ updates its model by
\begin{equation}
    \label{update_rule_pasgd}
    \bm{\theta}_i^{t,s+1} = \bm{\theta}_i^{t,s} -\eta g(\bm{\theta}_i^{t,s}),
\end{equation}
where $g(\bm{\theta}_i^{t,s}):= (1/\gamma)\sum_{\xi_i\in X_i}\nabla l(\bm{\theta}_i^{t,s}, \xi_i)$ represents the stochastic gradient computed based on a mini-batch $X_i$ of {$\gamma$} datapoints sampled from the local dataset ${D}_i$. Note that when $s=0$, the local model $\bm{\theta}_i^{t,s} = \bm{\theta}^{t}$ for all devices in $\Omega_t$. After $\tau$ local iterations, the selected devices upload their local models to the server where the shared model is updated by $\bm{\theta}^{t+1} = (1/r)\sum_{i\in\Omega_{t}} \bm{\theta}_i^{t,\tau}$. Therefore, each device is selected to participate with probability $r/n$ in each round and only needs to periodically communicate for $rT/n$ rounds in expectation. 
  

\subsection{Threat Model}\label{subsec:threat_model}
The adversary considered here can be the ``honest-but-curious'' central server or devices in the system. The central server will honestly follow the designed training protocol, but are curious about a target device's private data and may infer it from the shared messages. Furthermore, some devices could collude with the central server or each other to infer private information about the target device. Besides, the adversary can also be the passive outside attacker. These attackers can eavesdrop all shared messages in the execution of the training protocol but will not actively inject false messages into or interrupt the message transmission. Malicious devices who, for instance, may launch data pollution attacks by lying about their private datasets or returning incorrect computed results to disrupt the training process are out of the scope of this paper and will be left as our future work. 

\section{Our Private Federated Learning Scheme}\label{sec:DP-Fed}
In this section, we propose our method that enables multiple devices to jointly learn an accurate model for a given machine learning task in a private manner, without sacrificing much accuracy of the trained model. We first discuss how to preserve the data privacy of each device in the system with DP techniques. Then, we improve the accuracy of our method with secure aggregation and finally summarize the overall algorithm.



\subsection{Preventing Privacy Leakage with Differential Privacy}\label{subsec:private}

The aforementioned FedAvg method is able to prevent the direct information leakage of devices via keeping the raw data locally, however, it could not prevent more advanced attacks that infer private information of local training data by observing the messages communicated between devices and the server \cite{al2016reconstruction,shokri2017membership}. 
According to our threat model described in Section~\ref{subsec:threat_model}, devices and the server in the system are ``honest-but-curious'', and attackers outside the system can eavesdrop the transmitted messages. These attackers are able to obtain the latest shared model $\bm{\theta}^{t}$ sent from the server to devices and the local models $\{\bm{\theta}_i^{t,\tau}\}_{i\in\Omega_t}$ sent from devices to the server, both of which contain the private information of devices' training data. Our goal is to prevent the privacy leakage from these two types of messages with DP techniques. 


Towards that goal, we leverage the gradient perturbation with Gaussian noise \cite{abadi2016deep} to achieve DP so that the attacker is not able to learn much about an individual sample in ${D}_i$ from the shared massages. Specifically, at $s$-th local iteration of $t$-th round, device $i\in\Omega_t$ updates its local model by
\begin{equation}\label{update_rule_dppasgd}
\bm{\theta}_i^{t,s+1} = \bm{\theta}_i^{t,s} -\eta \left(g(\bm{\theta}_i^{t,s}) + \mathbf{b}_i^{t,s}\right),
\end{equation}
where $\mathbf{b}_i^{t,s}$ is the Gaussian noise sampled from the distribution $\mathcal{N}(0, \sigma^2\mathbf{I}_d)$. After $\tau$ local iterations, the uploaded local model $\bm{\theta}_i^{t,\tau} $ will preserve a certain level of DP guarantee for device $i$, which is proportional to the size of noise $\sigma$. Due to the post-processing property of DP\cite{dwork2014algorithmic}, the sum of local models, i.e., the updated shared model $\bm{\theta}^{t+1}$, will also preserve the same level of DP guarantee for device $i$. 

\subsection{Improving Model Utility with Secure Aggregation}

Although DP can be achieved using the above mechanism, the accuracy of the learned model may be low due to the large noise magnitude. At each round of the training, all uploaded local models are exposed to the attacker, leading to a large amount of information leakage. However, we observe that the server only needs to know the average values of the local models. Intuitively, one can reduce the privacy loss of devices by hiding the individual local models and restricting the server to receive only the sum of local models without disturbing the learning process. This can be achieved via a secure aggregation protocol so that the server can only decrypt the sum of the encrypted local models of selected devices without knowing each device's local model. In the following, we follow the principle of \cite{bonawitz2017practical} and design a customized protocol in our setting, which is efficient in terms of the amortized computation and communication overhead across all communication rounds. Note that one of the main contributions in this paper is to analyze the benefits of secure aggregation in reducing privacy loss as elaborated in Section~\ref{sec:privacy_analysis} rather than optimizing the design of the secure aggregation protocol.  



In our setting, a secure aggregation protocol should be able to 1) hide individual messages for devices, 2) recover the sum of individual messages of a random set of devices at each round, and 3) incur low communication cost for participating devices. Denote by $p_i^t$ the plaintext message of device $i$ that needs to be uploaded to the server. Note that the secure aggregation protocol only works for integers, hence the local model parameter $\bm{\theta}_i^{t,\tau}$ should be converted optimally to integers to obtain the plaintext $p_i^t$ \cite{bonawitz2016practical}. 
Our proposed protocol consists of the following two main steps: 
\begin{itemize}
\item \emph{Encryption uploading:} Devices in $\Omega_t$ upload their own encrypted local models $\{c_i^t\}_{i \in \Omega_t}$ to the server.
\item \emph{Decryption:} The server decrypts the sum of the messages received from devices in $\Omega_t$. 
\end{itemize}

\begin{figure}[t] 
\centering
\includegraphics[width=3in]{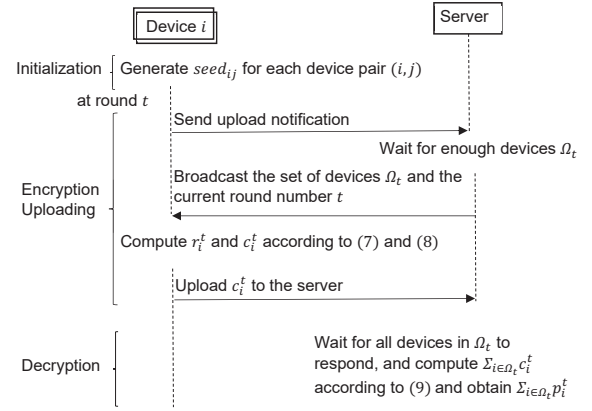}
\caption{Basic protocol for efficient secure aggregation in our approach.} 
\label{fig:protocol1}
\vspace*{-10pt}
\end{figure}

The basic idea of the protocol is to protect the message $p_i^t$ of device $i$ by hiding it with a random number $r_i^t$ in the plaintext space, i.e., $c_i^t = p_i^t + r_i^t$. 
However, the challenge here is how to remove the random number $r_i^t$ from the received ciphertext at the server part. To this end, we require that $\sum_{i \in \Omega_t} r_i^t = 0$, which prevents the attacker from recovering each individual message $p_i^t$ but enables the server to recover $\sum_{i\in \Omega_t} p_i^t$. However, this requires the devices to communicate with each other in order to generate such secrets $\{r_i^t\}_{i \in \Omega_t}$, which is inefficient in terms of communication overhead. 

To save the communication overhead, we introduce a pseudorandom function (PRF) $G$ here. The PRF $G$ takes a random seed $seed_{i,j}$ that both device $i$ and $j$ agree on during initialization and the round number $t$, and outputs a different pseudorandom number $G(seed_{i,j},t)$ at each round. Device $i$ could calculate the shared secret $r_{ij}^t$ without interacting with device $j$ at each round as long as they both use the same seed and round number, and thus each device could calculate $r_i^t$ without interactions. This procedure greatly reduces the amortized communication overhead of our protocol over multiple rounds.  

The detailed protocol is depicted in Fig.~\ref{fig:protocol1}. All devices need to go through an initialization step upon enrollment which involves pairwise communications with all other devices (which can be facilitated by the server) to generate a random seed $seed_{ij}$. After this initialization step, all enrolled devices could upload their messages through the encryption uploading step. At each round, only a subset of selected devices would upload their messages. Devices send a notification signal to the server once they are ready to upload their local models, and the server waits until receiving notifications from enough devices. The server then broadcasts the information $\Omega_t$ to all devices in $\Omega_t$. Device $i \in \Omega_t$ would first compute its secret at the current round as follows:
\begin{equation}
r_i^t =  \sum_{j\in \Omega_t\setminus\{i\}} \left(r_{ij}^t-r_{ji}^t\right),
\end{equation}
where $r_{ij}^t = G(seed_{i,j},t)$ is a secret known by both device $i$ and $j$. Device $i$ could then generate the ciphertext for $p_i^t$ by
\begin{equation}
    c_i^t = p_i^t + r_i^t.
\end{equation}
In the decryption step, the server receives $\{c_i^t\}_{i\in\Omega_t}$ from all selected devices. The server could then recover the sum of plaintext messages from devices in $\Omega_t$ as follows:
\begin{equation}
\begin{split}
\sum_{i\in \Omega_t} c_i^t & = \sum_{i\in \Omega_t} p_i^t + \sum_{i\in \Omega_t}\sum_{j\in \Omega_t\setminus\{i\}}(r_{ij}^t-r_{ji}^t) \\
& = \sum_{i \in \Omega_t} p_i^t.
\end{split}
\end{equation}
Note that in the above protocol, we assume all devices in $\Omega_t$ have stable connection to the server. 
The entire process of our approach, integrating gradient perturbation, secure aggregation, and multiple steps of local SGD, is summarized in Algorithm~\ref{algorithm-1}. 


\begin{algorithm}[ht]
\caption{Private Federated Learning Algorithm}
\label{algorithm-1}
\begin{algorithmic}[1]
\REQUIRE number of rounds $T$, local iteration period $\tau$, number of selected devices per round $r$, stepsize $\eta$
\FOR{$t=0$ to $T-1$}
    \STATE Server uniformly selects a set $\Omega_t$ of $r$ devices
    \STATE Server broadcasts $\bm{\theta}^t$ to all devices in $\Omega_t$
    \FOR{all devices in $\Omega_t$ in parallel}
        \STATE $\bm{\theta}_i^{t,0}\leftarrow \bm{\theta}^t$
        \FOR{$s=0$ to $\tau-1$}
            \STATE Sample a mini-batch $X_i$ and compute gradient $g(\bm{\theta}_i^{t,s}) \gets (1/\gamma)\sum_{\xi_i\in X_i}\nabla l(\bm{\theta}_i^{t,s},\xi_i)$
            \STATE Sample DP noise $\bm{b}_{i}^{t, s} \sim \mathcal{N}(0, \sigma^2 \mathbf{I}_d)$
            \STATE $\bm{\theta}_i^{t,s+1} \leftarrow \bm{\theta}_i^{t,s} - \eta \left(g(\bm{\theta}_i^{t,s}) + \bm{b}_{i}^{t, s} \right)$
       \ENDFOR
        \STATE Generate encrypted local model $c_i^{t}$ using the secure aggregation protocol and send it to the server
    \ENDFOR
    \STATE Server decrypts the average of the received local models $(1/r)\sum_{i\in\Omega_t} {c}_i^{t}$ to get the new global model $\bm{\theta}^{t + 1}$
    \ENDFOR
\end{algorithmic}
\end{algorithm}

\section{Privacy Analysis}\label{sec:privacy_analysis}

As mentioned before, our goal of using DP techniques is to prevent the outside attacker or the ``honest-but-curious'' server and devices from learning sensitive information about the local data of a device. Using the secure aggregation protocol, the local model is encrypted and the attacker can only obtain the sum of local models. Thus, as long as the sum of local models is differentially private, we can prevent the attacks launched by the attacker. 

Instead of using the traditional $(\epsilon, \delta)$-DP notion, we use zCDP to tightly account the end-to-end privacy loss of our approach across multiple iterations and then convert it to an $(\epsilon, \delta)$-DP guarantee. 
%
In the following, we first compute the sensitivity of the gradient $g(\bm{\theta}_i^{t,s})$ (as given in Corollary~\ref{colo:sensitivity_g}) based on Assumption~\ref{assp:local_unbiased} to show that each iteration of Algorithm~\ref{algorithm-1} achieves zCDP. Note that Assumption~\ref{assp:local_unbiased} is a common assumption for differentially private learning algorithms and can be achieved by gradient clipping techniques\cite{abadi2016deep}. Then, we compute the sensitivity of the uploaded local model $ \bm{\theta}_i^{t,\tau}$ (as given in Lemma~\ref{lema:sensitivity_local}) to further capture the zCDP guarantee of each communication round. Finally, we show that Algorithm~\ref{algorithm-1} satisfies $(\epsilon_i,\delta)$-DP for device $i$ after $T$ communication rounds in Theorem~\ref{thm:privacy}.

\begin{assumption}[Bounded gradients]
\label{assp:local_unbiased}
The $L_2$-norm of the stochastic gradient $\nabla l(\mathbf{x}, \xi)$ is bounded, i.e.,  for any $ \mathbf{x}\in\mathbb{R}^d$ and $\xi \in \bigcup_{i \in [n]} D_i$, $\|\nabla l(\mathbf{x},\xi)\|_2 \leq G$.
\end{assumption}
\begin{corollary}
\label{colo:sensitivity_g}
The sensitivity of the stochastic gradient $g(\bm{\theta}_i^{t,s})$ of device $i$ at each local iteration  is bounded by ${2G}/{\gamma}$.
\end{corollary}
\begin{IEEEproof}
The proof of Corollary~\ref{colo:sensitivity_g} is provided in Appendix~\ref{proof:sensitivity_g}.
\end{IEEEproof}
By Lemma~\ref{rho-zcdp} and Corollary~\ref{colo:sensitivity_g}, each iteration of Algorithm~\ref{algorithm-1} achieves $ {2 G^2}/{\gamma^2\sigma^2}$-zCDP for every active device at this iteration. During the local computation, the local dataset will be randomly shuffled and partitioned into $m/\gamma$ mini-batches, each containing $\gamma$ datapoints. Assume $\tau$ is divided evenly by $m/\gamma$, then the whole local dataset will be accessed for $\tau \gamma/m$ times at each round. Therefore, at round $t$, the uploaded local model $\bm{\theta}_i^{t,\tau}$ satisfies $ {2\tau G^2}/{m\gamma\sigma^2}$-zCDP by using Lemma~\ref{zcdp-shuffling} and Lemma~\ref{composition}.

Given the zCDP guarantee of $\bm{\theta}_i^{t,\tau}$ and its sensitivity given in Lemma~\ref{lema:sensitivity_local}, we can observe that the variance of Gaussian noise added to $\bm{\theta}_i^{t,\tau}$ is equivalent to $m\tau\eta^2\sigma^2/\gamma$. Therefore, we can obtain that the variance of Gaussian noise added to the sum of uploaded local models is $rm\tau\eta^2\sigma^2/\gamma$, due to the independence of Gaussian noise. By Lemma~\ref{rho-zcdp}, we can obtain the zCDP guarantee of the sum of uploaded local models if we can measure the sensitivity of $\sum_{i\in\Omega_t} \bm{\theta}_i^{t,\tau}$. It is easy to verify that, for device $i\in\Omega_t$, the sensitivity of the sum of uploaded local models $\sum_{i\in\Omega_t} \bm{\theta}_i^{t,\tau}$ is equivalent to the sensitivity of $\bm{\theta}_i^{t,\tau}$. Finally, we obtain that $\sum_{i\in\Omega_t} \bm{\theta}_i^{t,\tau}$ satisfies $ 2\tau G^2/rm\gamma \sigma^2$-zCDP, which means round $t$ of Algorithm~\ref{algorithm-1} achieves $ {2\tau G^2}/{r  m\gamma\sigma^2}$-zCDP for each device in $\Omega_t$. Finally, we compute the overall privacy guarantee for a device after $T$ communication rounds and give the $(\epsilon,\delta)$-DP guarantee in Theorem~\ref{thm:privacy}.

\begin{lemma}
\label{lema:sensitivity_local}
The sensitivity of the uploaded local model $ \bm{\theta}_i^{t,\tau}$ at round $t$ is bounded by ${2\eta \tau G }/\gamma$.
\end{lemma}
\begin{IEEEproof}
The proof of Lemma~\ref{lema:sensitivity_local} is provided in Appendix~\ref{proof:sensitivity_local}.
\end{IEEEproof}

\begin{theorem}
\label{thm:privacy}
{In Algorithm~\ref{algorithm-1},} let the mini-batch $X_i$ be randomly sampled without replacement from $D_i$ every $m/\gamma$ local iterations and the Gaussian noise $\mathbf{b}_i^{t,s}$ be sampled from $ \mathcal{N}(0,\sigma^2\mathbf{I}_d)$. Assume $\tau$ can be divided evenly by $m/\gamma$, and let $C_i$ represent the number of rounds device $i$ gets selected for out of $T$ communication rounds, then Algorithm~\ref{algorithm-1} achieves $(\epsilon_i,\delta)$-DP for device $i$ in the system, where
\begin{equation}
    \epsilon_i = \frac{2C_i\tau G^2}{r m\gamma \sigma^2} + 2\sqrt{\frac{2C_i\tau G^2}{r m\gamma \sigma^2} \log{\frac{1}{\delta}} }.
\end{equation}
\end{theorem}
\begin{IEEEproof}
The proof of Theorem~\ref{thm:privacy} is provided in Appendix~\ref{proof:privacy}.
\end{IEEEproof}

\section{Convergence Analysis}\label{sec:convg}
In this section, we present the main theoretical results on the convergence properties of our approach. 
Before stating our results, we give some assumptions and summarize the update rule of our approach as follows. Assumption~\ref{assp:smooth} implies that the objective function $f$ are $L$-smooth. Assumption~\ref{assp:bounded_divergence} ensures that the divergence between local stochastic gradients is bounded. These two assumptions are standard in literature\cite{li2019convergence,wang2018cooperative,stich2018local}. 

\begin{assumption}[Smoothness]
\label{assp:smooth}
The local objective function $f_i$ is $L$-smooth, i.e., for any $\mathbf{x}, \mathbf{y}\in \mathbb{R}^d$ and $i\in[n]$, we have $f_i(\mathbf{y}) \leq f_i(\mathbf{x}) + \langle\nabla f_i(\mathbf{x}), \mathbf{y}-\mathbf{x}\rangle + ({L}/{2})\|\mathbf{y}-\mathbf{x}\|^2$.
\end{assumption}
\begin{assumption}[Bounded divergence]
\label{assp:bounded_divergence}
Let $\xi_i$ be randomly sampled from the local dataset $D_i$. The stochastic gradient of each device is unbiased and will not diverge a lot from the exact gradient, i.e., for any $\mathbf{x}\in\mathbb{R}^d$ and $i\in[n]$, $\mathbb{E}[\nabla l (\mathbf{x}, \xi_i)] = \nabla f_i(\mathbf{x})$ and $ \mathbb{E}\|\nabla l (\mathbf{x}, \xi_i) - \nabla f_i(\mathbf{x})\|^2 \leq \beta^2$.
\end{assumption}


To prove the convergence of our approach, we first represent the update rule of our approach in a general manner. In Algorithm~\ref{algorithm-1}, the total number of iterations is $K$, i.e., $K=T\tau$. 
At iteration $k$ where $k=t\tau + s$, each device $i$ evaluates the stochastic gradient $g(\bm{\theta}_i^k)$ based on its local dataset and updates current model $\bm{\theta}_i^k$. Thus, $n$ devices have different versions $\bm{\theta}_1^k,\dots,\bm{\theta}_n^k $ of the model. After $\tau$ local iterations, devices upload their encrypted local models to the server to generate the new shared model, i.e., $({1}/{r})\sum_{i\in\Omega_k}{\bm{\theta}_i^k}$ with $(k\mod \tau=0)$, where $\Omega_{k} =\Omega_{t}, \forall k \in [t \tau, t \tau + 1, \ldots, t \tau + \tau - 1]$.

Now, we can present a virtual update rule that captures Algorithm~\ref{algorithm-1}. Define matrices $\bm{\Theta}^k, \mathbf{G}^k, \mathbf{B}^k \in \mathbb{R}^{d\times n}$ for $k=0,\dots,K-1$ that concatenate all local models, gradients and noises:
\begin{equation*}
\begin{split}
   & \bm{\Theta}^k := \left[\bm{\theta}_1^{k}, \bm{\theta}_2^{k},  \dots, \bm{\theta}_n^{k} \right],\\
   & \mathbf{G}^k := \left[ g(\bm{\theta}_1^{k}),g(\bm{\theta}_2^{k}), \dots, g(\bm{\theta}_n^{k})\right],\\
   & { \mathbf{B}^k := \left[\mathbf{b}_1^k,\mathbf{b}_2^k, \dots, \mathbf{b}_n^k\right]}.
\end{split}
\end{equation*}
If device $i$ is not selected to upload its model at iteration $k$, $\bm{\theta}_i^{k} =g(\bm{\theta}_i^{k})=\mathbf{b}_i^k=\mathbf{0}_d$. Besides, define matrix $\mathbf{J}^{\Omega_k} \in\mathbb{R}^{n\times n}$ with element $\mathbf{J}^{\Omega_k}_{i,j}={1}/{r}$ if $i\in\Omega_k$ and $\mathbf{J}^{\Omega_k}_{i,j}=0$ otherwise. Unless otherwise stated, $\mathbf{1}^k \in \mathbb{R}^{n}$ is a column vector of size $n$ with element $\mathbf{1}_{i}^k = 1 $ if $i\in\Omega_k$ and $\mathbf{1}_{i}^k=0$ otherwise. To capture periodic averaging, we define $\mathbf{J}^k$ as
\begin{equation*}
\mathbf{J}^k:= 
\begin{cases}
{ \mathbf{J}^{\Omega_k}}, k \text{ mod } \tau =0\\
\mathbf{I}_{n}, \text{ otherwise.}
\end{cases}
\end{equation*}
where $\mathbf{I}_{n}$ is a $n\times n$ identity matrix. Then a general update rule of our approach can be expressed as follows:
\begin{equation}
\label{averaged_model}
\bm{\Theta}^{k+1} =\left(\bm{\Theta}^{k} - \eta (\mathbf{G}^k + { \mathbf{B}^k)}\right) \mathbf{J}^k.
\end{equation}
Note that the secure aggregation does not change the sum of local models. Multiplying $\textbf{1}^k/r$ on both sides of \eqref{averaged_model}, we have
\begin{equation}
\label{avg}
\frac{\bm{\Theta}^{k+1}\mathbf{1}^k}{r}=\frac{\bm{\Theta}^{k}\mathbf{1}^k}{r} - \eta \left( \frac{ \mathbf{G}^k\mathbf{1}^k}{r} + { \frac{ \mathbf{B}^k\mathbf{1}^k}{r}}\right).
\end{equation}
Then define the averaged model at iteration $k$ as
\begin{equation*}
\hat{\bm{\theta}}^k := \frac{\bm{\Theta}^{k}\mathbf{1}^k}{r} = \frac{1}{r} \sum_{i\in\Omega_k} \bm{\theta}_i^{k}. 
\end{equation*}
After rewriting \eqref{avg}, one yields
\begin{equation}
\hat{\bm{\theta}}^{k+1} = \hat{\bm{\theta}}^k - \eta \left(\frac{1}{r} \sum_{i\in\Omega_k} g(\bm{\theta}_i^{k}) +{  \mathbf{b}_i^k}\right).
\end{equation}

Since devices are selected at random to perform updating in each round, and $g(\bm{\theta}_i^{k})$ is the stochastic gradient computed on a subset of data samples $X_i\in D_i$, the randomness in our federated learning system comes from the device selection, stochastic gradient, and Gaussian noise. In the following, we bound the expectation of several intermediate random variables, 
which we denote by $\mathbb{E}_{\{\Omega_k, X_i, \mathbf{b}_i^k|i\in[n]\}}[\cdot]$. {To simplify the notation}, we use $\mathbb{E}[\cdot]$ instead of $\mathbb{E}_{\{\Omega_k, X_i, \mathbf{b}_i^k|i\in[n]\}}[\cdot]$ in the rest of the paper, unless otherwise stated. 
 



\begin{theorem}[Convergence Result of Our Approach]
\label{thrm:non-convex}
For Algorithm~\ref{algorithm-1}, suppose the total number of iterations $K=T\tau$ where $T$ is the number of communication rounds and $\tau$ is the local iteration period. Under Assumptions \ref{assp:smooth}-\ref{assp:bounded_divergence}, if the learning rate satisfies $5\eta L + 3\tau^2\eta^2 L^2\leq 1$, and all devices are initialized at the same point $\bm{\theta}^0 \in \mathbb{R}^d$, then after $K$ iterations the expected gradient norm is bounded as follows
\begin{multline}
   \mathbb{E}\left[\frac{1}{K}\sum_{k=0}^{K-1} \|\nabla f(\hat{\bm{\theta}}^{k})\|^2 \right] 
   \leq \frac{2(f(\bm{\theta}^{0}) -f^*)}{\eta K} + \frac{\eta L\beta^2}{2\gamma} + \frac{\eta Ld\sigma^2}{2r}\\
    + (\tau-1)(2\tau-1)\left(\frac{2\eta^2 L^2\beta^2}{3\gamma} + \frac{\eta^2 L^2d\sigma^2(r+1)}{3r}\right).
\end{multline}
Here, $\sigma^2$ is the variance of Gaussian noise, {$\beta^2$ is the upper bound of the variance of local stochastic gradients}, $L$ is the Lipschitz constant of the gradient, $n$ is the total number of devices, and $r$ is the number of selected devices at each round. 
\end{theorem}
\begin{IEEEproof}
The proof of Theorem~\ref{thrm:non-convex} is provided in Appendix~\ref{proof:non}.
\end{IEEEproof}

By setting the learning rate $\eta=\mathcal{O}(\sqrt{{n}/{K}})$, Algorithm~\ref{algorithm-1} achieves the asymptotic convergence rate of $\mathcal{O}(1/\sqrt{nK}) + \mathcal{O}( {\tau^2\sigma^2}/{K})$, when $K$ is sufficiently large. If we further assume that the objective function is strongly convex, i.e., the following Assumption~\ref{assp:convex} holds, Algorithm~\ref{algorithm-1} achieves the non-asymptotic convergence result stated in Lemma~\ref{thrm:convex}. 
\begin{assumption}
\label{assp:convex}
The objective function $f$ is $\lambda$-strongly convex if for any $\mathbf{x},\mathbf{y}\in\mathbb{R}^d$ we have $ \|\nabla f(\mathbf{x}) -\nabla f(\mathbf{y}) \| \geq \lambda\|\mathbf{x}-\mathbf{y}\|$ for some constant $\lambda > 0$.
\end{assumption}


\begin{lemma}[Convergence Result for Convex Loss] \label{thrm:convex}
For Algorithm~\ref{algorithm-1}, suppose the total number of iterations $K=T\tau$ where $T$ is the number of communication rounds and $\tau$ is the local iteration period. Under Assumptions \ref{assp:smooth}-\ref{assp:convex}, if the learning rate satisfies $5\eta L + 3\tau^2\eta^2 L^2\leq 1$, and all devices are initialized at the same point $\bm{\theta}^0 \in \mathbb{R}^d$. Then after $K$ iterations, the expected optimality gap is bounded as follows
\begin{multline}\label{conv_bound}
\mathbb{E}\left[\frac{1}{K}\sum_{k=0}^{K-1}f(\hat{\bm{\theta}}^k) - f^*\right]  \leq  \frac{(1-{\eta \lambda})}{K{\eta \lambda}} \left(f({\bm{\theta}}^{0}) - f^*\right) +\frac{\eta L\beta^2}{4\lambda \gamma} \\
 +\frac{\eta Ld\sigma^2}{4r\lambda} + \eta^2 L^2(\tau-1)(2\tau-1)\left( \frac{\beta^2}{3\lambda \gamma} + \frac{d\sigma^2(r+1)}{6r\lambda}\right).
\end{multline}
\end{lemma}
\begin{IEEEproof}
The proof of Lemma~\ref{thrm:convex} is provided in Appendix~\ref{proof:convex}.
\end{IEEEproof}


\section{Experiments}\label{sec:exp}
In this section, we evaluate the performance of our proposed scheme. We first describe our experimental setup and then show the convergence properties of our approach. Next, we demonstrate the effectiveness of our approach by comparing it with a baseline approach. Finally, we show the trade-off between privacy and model utility in our approach and how our secure aggregation protocol improves the accuracy of the learned model.

\begin{figure*}[!htb]
\centering
\includegraphics[width=0.9\linewidth]{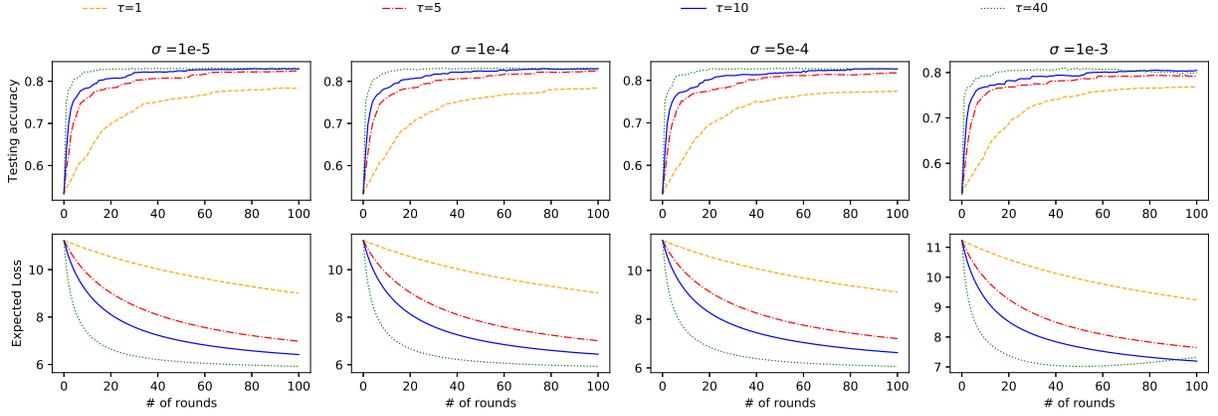}
\caption{Convergence of the expected loss (logistic regression). Here, we show the convergence of the first 100 communication rounds.}
\label{fig:converge_lr}
\vspace*{-10pt}
\end{figure*}
\begin{figure*}[!htb]
\centering
\includegraphics[width=0.9\linewidth]{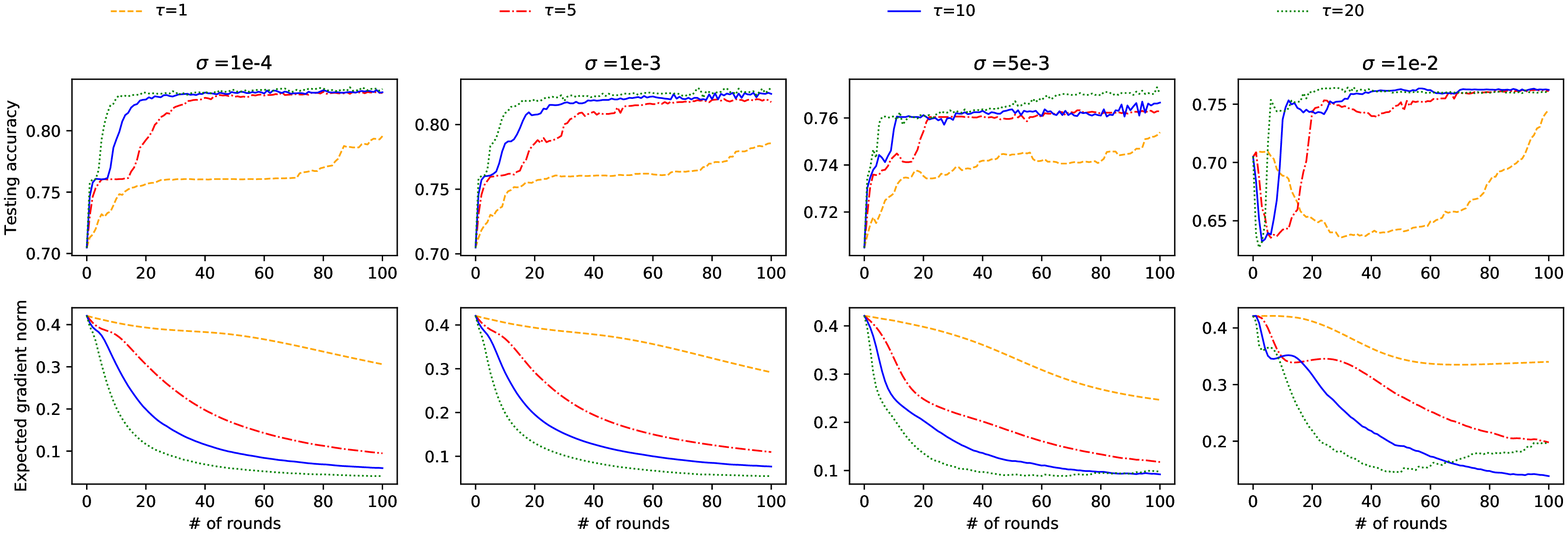}
\caption{Convergence of the expected gradient norm (neural network). Here, we show the convergence of the first 100 communication rounds.}
\label{fig:converge_mlp}
\vspace*{-10pt}
\end{figure*}

\subsection{Experimental Setup}\label{subsec:exp_setup}

\textbf{Datasets and Learning Tasks.} We explore the benchmark dataset \emph{Adult}\cite{blake1998uci} using both logistic regression and neural network models in our experiments. 
The Adult dataset contains 48,842 samples with 14 numerical and categorical features, with each sample corresponding to a person. The task is to predict if the person's income exceeds $\$50,000$ based on the 14 attributes, namely, \textit{age, workclass, fnlwgt, education, education-num, marital-status, occupation, relationship, race, sex, capital-gain, capital-loss, hours-per-week, and native-country}.
To simulate a distributed setting based on the Adult dataset, we evenly assign the original Adult data to 16 devices such that each device contains 3,052 data samples. 
We train a logistic regression classifier and a 3-layer neural network classifier (with ReLU activation function) and use the softmax cross-entropy as the loss function. 

\textbf{Baseline.} We select the state-of-the-art differentially private learning scheme named DP-SGD \cite{abadi2016deep} as a baseline to evaluate the efficiency of our proposed scheme. In DP-SGD, only one step of SGD is performed to update the local model on a device during each communication period, and Gaussian noise is added to each model update before sending it out. 

\textbf{Hyperparameters.} We take $80\%$ of the data on each device for training, $10\%$ for testing and $10\%$ for validation. We tune the hyperparameters on the validation set and report the average accuracy on the testing sets of all devices. The gradient norm $G$ is enforced by clipping, which is widely used in differentially private learning. For all experiments, we set the privacy failure probability $\delta = 10^{-4}$ and the number of selected devices per round $r=10$. Note that due to the randomized nature of differentially private mechanisms, we repeat all the experiments for 5 times and report the average results.

\subsection{Convergence Properties of Our Approach}
In this subsection, we show the algorithmic convergence properties of our approach under several settings of noise magnitude $\sigma$ and local iteration period $\tau$. Specifically, for the logistic regression, we show the testing accuracy and the expected training loss 
with respect to the number of communication rounds $T$ when $\sigma \in \{10^{-5},10^{-4},5\times 10^{-4},10^{-3}\}$ and $\tau \in \{1,5,10,40\}$. The results for the logistic regression are depicted in Fig.~\ref{fig:converge_lr}.
Similarly, for the neural network, we show the testing accuracy and expected gradient norm 
with respect to the number of communication rounds $T$ when $\sigma \in \{10^{-4},10^{-3},5\times 10^{-3},10^{-2}\}$ and $\tau \in \{1,5,10,20\}$. The results for the neural network are finally shown in Fig.~\ref{fig:converge_mlp}.

For the logistic regression, the testing accuracy and expected loss will generally decrease sharply and then slowly afterwards. 
As the noise magnitude $\sigma$ increases, the expected training loss of the logistic regression converges to a higher bound and the testing accuracy decreases, which is consistent with the convergence properties of our approach where larger $\sigma$ implies larger convergence error.
For all settings of noise, with larger local iteration period, the expected loss drops more sharply at the beginning and arrives at a higher stationary point, which is consistent with our approach's convergence properties where larger $\tau$ implies larger convergence error.
When $\sigma=10^{-3}$ and $\tau=40$, we can see that after the expected loss decreases to 7 using about 40 rounds of communication, it increases as more computations and communications are involved. The reason is that after the loss arrived at a stationary point, keeping training brings additional noise into the well-trained model and hence the model performance drops.
Similar trends have been observed for the neural network classifier. When $\sigma=10^{-2}$, the testing accuracy drops from the initialized value quickly as the noise is added into the system, and then it increases as more computations and communications are involved. 

\subsection{Model Utility of Our Approach}

\begin{figure}[t]
\centering
\includegraphics[width=0.9\linewidth]{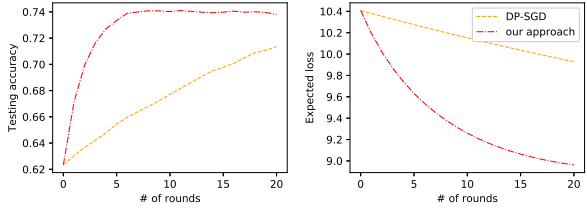}
\caption{Testing accuracy and expected training loss of our approach in comparison with DP-SGD (logistic regression). Here, we set $T=20$ and $\epsilon=10$ for both approaches, and set $\tau=10$ for our approach.}
\label{fig:communicate_mlp}
\vspace*{-10pt}
\end{figure}

\begin{figure}[t]
\centering
\includegraphics[width=0.9\linewidth]{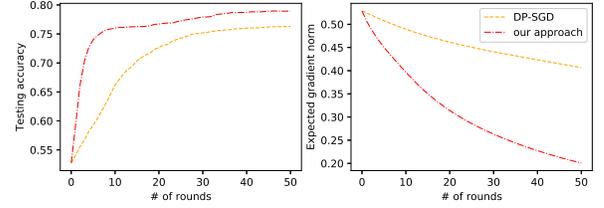}
\caption{Testing accuracy and expected gradient norm of our approach in comparison with DP-SGD (neural network). Here, we set $T=50$ and $\epsilon=10$ for both approaches, and set $\tau=5$ for our approach.}
\label{fig:communicate_lr}
\vspace*{-10pt}
\end{figure}

In this subsection, we show the model utility of our approach compared with DP-SGD. Specifically, for the logistic regression, we set the number of communication rounds $T=20$ for both approaches and $\tau=10$ for our approach. Both approaches preserve $(10,10^{-4})$-DP after 20 rounds of communication. For our approach, we compute the noise magnitude $\sigma$ by Theorem~\ref{thm:privacy}. 
Note that we randomly sample the active devices for each round and make sure each device participates the same number of communication rounds beforehand, and then we use the sampling result for both approaches. 
For DP-SGD, it achieves $(2C_iG^2/n\gamma^2\sigma^2 + 2\sqrt{2\log(1/\delta)C_iG^2/n\gamma^2\sigma^2}, \delta)$-DP for device $i$, which can be used to compute the noise magnitude $\sigma$ given $\epsilon$ and $\delta$. 
The testing accuracy and expected loss with respect to the number of communicate rounds are shown in Fig.~\ref{fig:communicate_lr}. For the neural network, we set the number of communication rounds $T=50$, the overall privacy budget $\epsilon=10$ for both approaches, and $\tau=5$ for our approach. 
The testing accuracy and expected gradient norm with respect to the number of communication rounds are shown in Fig.~\ref{fig:communicate_mlp}. 
%
For logistic regression, we observe that our approach exhibits faster convergence than DP-SGD at the beginning, and finally achieves higher accuracy and lower expected loss than DP-SGD within 20 rounds of communication. For neural network, we observe the similar trend. Our approach converges faster than DP-SGD and achieves higher accuracy and lower expected gradient norm than DP-SGD. Therefore, our approach can achieve higher model utility than DP-SGD under the same privacy guarantee.

\subsection{Trade-off between Privacy and Utility}
\begin{figure}[t]
\centering
\includegraphics[width=0.9\linewidth]{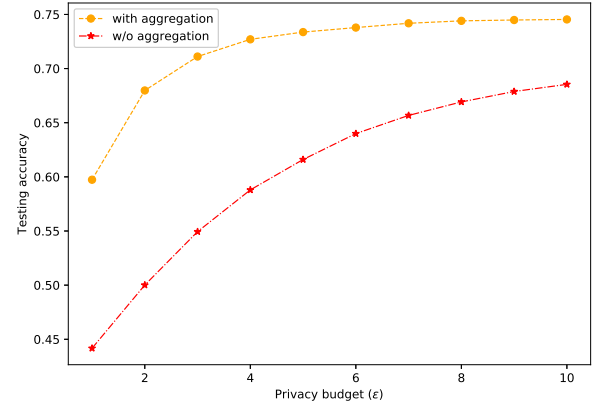}
\caption{Trade-off between privacy and accuracy (logistic regression). Here, we set $\tau=2$ and $ T=20$.}
\label{fig:tradeoff_lr}
\vspace*{-10pt}
\end{figure}

\begin{figure}[t]
\centering
\includegraphics[width=0.9\linewidth]{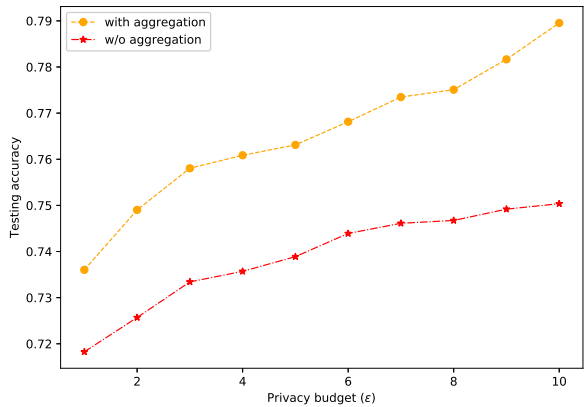}
\caption{Trade-off between privacy and accuracy (neural network). Here, we set $T=50$ and $\tau=5$.}
\label{fig:tradeoff_mlp}
\vspace*{-10pt}
\end{figure}

To observe the privacy-utility tradeoff, we evaluate the effects of different values of privacy budgets $\epsilon$ on the testing accuracy of trained classifiers. 
In addition, we compare our approach with the approach without secure aggregation (i.e., same as our approach but without secure aggregation) to show how secure aggregation improves the accuracy. 
For logistic regression, we set the local iteration period $\tau=2$ and the number of communication rounds $T=20$. For neural network, we set the local iteration period $\tau=5$ and the number of communication rounds $T=50$. We show the testing accuracy with respect to different values of privacy budget $\epsilon$ of logistic regression and neural network in Fig.~\ref{fig:tradeoff_lr} and Fig.~\ref{fig:tradeoff_mlp}, respectively. As expected, a larger value of $\epsilon$ results in higher accuracy while providing lower DP guarantee. Moreover, our approach with secure aggregation always outperforms the approach without secure aggregation because less noise is added in each iteration. 

\section{Related Work}\label{sec:related}
Privacy issue has received significant attention recently in distributed learning scenarios handling user-generated data. Among distributed learning schemes that preserve privacy, many of them rely on secure multi-party computation or homomorphic encryption, which involve both high computation and communication overhead and are only applicable to simple learning tasks such as linear regression\cite{NiWe13} and logistic regression\cite{MoZh17}. 
Furthermore, these privacy-preserving solutions could not prevent the information leakage from the final learned model. 
DP has become the de-facto standard for privacy notion and is being increasingly adopted in private data analysis \cite{dwork2014algorithmic}. 
A wide range of differentially private distributed learning algorithms (see \cite{guo2018practical,huang2019dp,abadi2016deep,huang2012differentially} and references therein) have been proposed based on different optimization methods (e.g., alternating direction method of multipliers, gradient descent, and distributed consensus) and noise addition mechanisms (e.g., output perturbation, objective perturbation, and gradient perturbation). 
However, most of them do not consider the communication efficiency aspect and therefore are not suitable for federated learning. 
Moreover, few of them could provide a rigorous performance guarantee. 

A few recent works \cite{agarwal2018cpsgd,li2019privacy,mcmahan2018learning,geyer2017differentially,wei2020federated} have started to consider the privacy aspects in federated learning. Specifically, Agarwal et al. \cite{agarwal2018cpsgd} proposed a modified distributed SGD scheme based on gradient quantization and binomial mechanism to make the scheme both private and communication-efficient. 
Li et al. \cite{li2019privacy} developed a method that compresses the transmitted messages via sketches to simultaneously achieve communication efficiency and DP in distributed learning. Our work is orthogonal to theirs by focusing on reducing the number of communication rounds via more local computation per round instead of the size of messages transmitted per round. McMahan et al. \cite{geyer2017differentially}, Geyer et al. \cite{mcmahan2018learning} and Liang et al. \cite{liang2020exploring} designed an approach to preserving DP in federated learning. However, their approaches assume a fully trusted server and do not provide any rigorous performance guarantee. Wei et al. \cite{wei2020federated} proposed a differentially private federated learning scheme assuming a trusted server and provided the performance guarantee for convex loss functions. In comparison, our scheme does not assume a trusted server and improves the model utility of federated learning with DP guarantee by using zCDP and secure aggregation. Furthermore, we provide a rigorous performance guarantee for both convex and non-convex loss functions.


\section{Conclusions}\label{sec:con}
This paper focuses on privacy-preserving federated learning. We have proposed a new federated learning scheme with rigorous privacy guarantee. Our proposed scheme preserves the DP of devices while achieving high accuracy of the resulting model. We have also provided rigorous convergence analysis of our proposed approach, and extensive experiments based on the real-world dataset have verified the effectiveness of the proposed scheme and shown the trade-off between model utility and privacy. For future work, we plan to study the performance of our approach in other learning settings such as multi-task learning and privacy considerations such as personalized DP. 

\ifCLASSOPTIONcaptionsoff
  \newpage
\fi
\bibliographystyle{IEEEtran}
\bibliography{hu,gong,guo}

\appendix

\subsection{Proof of Corollary~\ref{colo:sensitivity_g}\label{proof:sensitivity_g}}
\begin{corollary}
\label{colo:sensitivity_g}
The sensitivity of the stochastic gradient $g(\bm{\theta}_i^{t,s})$ of device $i$ at each local iteration  is bounded by ${2G}/{\gamma}$.
\end{corollary}
\begin{IEEEproof}
For device $i$, given any two neighboring datasets ${X}_i$ and ${X}_i^\prime$ of size $\gamma$ that differ only in the $j$-th data sample, the sensitivity of the stochastic gradient computed at each local iteration in Algorithm~\ref{algorithm-1} can be computed as
\begin{multline*}
\|g(\bm{\theta}_i^{t,s}; {X}_i) - g(\bm{\theta}_i^{t,s}; {X}_i^\prime)\|_2 \\
= \frac{1}{\gamma} \|\nabla l(\bm{\theta}_i^{t,s}; \xi_j ) - \nabla l(\bm{\theta}_i^{t,s}; \xi_j^\prime) \|_2. 
\end{multline*}
By Assumption~\ref{assp:local_unbiased}, the sensitivity of $g(\bm{\theta}_i^{t,s})$ can be estimated as $\Delta_2(g(\bm{\theta}_i^{t,s})) \leq 2G/\gamma$.
\end{IEEEproof}

\subsection{Proof of Lemma~\ref{lema:sensitivity_local}\label{proof:sensitivity_local}}

\begin{IEEEproof}
Without adding noise, the local model of device $i\in\Omega_t$ after $\tau$ local iterations at round $t$ can be written as
\begin{equation}
\label{eq:local_model_sequential}
    \bm{\theta}_i^{t,\tau} = \bm{\theta}_i^{t,0} -\eta g(\bm{\theta}_i^{t,0}) - \dots - \eta g(\bm{\theta}_i^{t,\tau-1}).
\end{equation}
According to the sensitivity of $g(\bm{\theta}_i^{t,s})$ given in Corollary~\ref{colo:sensitivity_g}, we have that \begin{align*}
    \Delta_2(\bm{\theta}_i^{t,\tau}) &= \eta \Big\| g(\bm{\theta}_i^{t,0}; {X}_i^{t,0}) - g(\bm{\theta}_i^{t,0}; {{X_i^{t,0}}^\prime}) + \dots 
   \\
   & \ \ + g(\bm{\theta}_i^{t,\tau-1}; {X}_i^{t,\tau-1}) - g(\bm{\theta}_i^{t,\tau-1}; {{X}_i^{t,\tau-1}}^\prime) \Big\|\\
   & \leq \frac{2\eta \tau G }{\gamma}.
\end{align*}
\end{IEEEproof}

\subsection{Proof of Theorem~\ref{thm:privacy}\label{proof:privacy}}

\begin{IEEEproof}
It is proved that each round of Algorithm~\ref{algorithm-1} achieves $ {2\tau G^2}/{r  m\gamma\sigma^2}$-zCDP for the device in $\Omega_t$. Due to the device selection, not all devices will upload their models to the server at round $t$. If their models are not sent out, they do not lose their privacy at that round. Let $C_i$ represent the number of communication rounds device $i$ participated during the whole training process. by Lemma~\ref{composition}, the overall zCDP guarantee of device $i$ in the system after $T$ rounds of communication is $ {2\tau C_i G^2}/{r  m\gamma\sigma^2}$. Therefore, Theorem~\ref{thm:privacy} then follows by Lemma~\ref{zcdp-dp}. Note that, each device in the system participates the communication with probability $r/n$ at each round, hence $C_i$ is equivalent to $Tr/n$ in expectation.
\end{IEEEproof}

\subsection{Proof of Theorem~\ref{thrm:non-convex}\label{proof:non}}
As given in Lemma~\ref{lema:unbiased_update} and Lemma~\ref{lema:network_error}, we compute the upper bound of the expectation of the perturbed stochastic gradients and the network error that captures the divergence between local models and the averaged model. Note that the proofs of Lemma~\ref{lema:unbiased_update} and Lemma~\ref{lema:network_error} are given in Section~\ref{proof:unbiased_update} and \ref{proof:network_error}. 
\begin{lemma}
\label{lema:unbiased_update}
The expectation and variance of the averaged perturbed stochastic gradients at iteration $k$ are
\begin{equation}
    \mathbb{E}\left[\frac{1}{r}\sum_{i\in\Omega_k} \left(g(\bm{\theta}_i^{k}) +{  \mathbf{b}_i^k}\right)\right] = \frac{1}{n}\sum_{i=1}^{n}\nabla f_i(\bm{\theta}_i^{k}),
\end{equation}
and
\begin{align}
    \nonumber & \mathbb{E}\left[\left\|\frac{1}{r}\sum_{i\in\Omega_k} \left(g(\bm{\theta}_i^{k}) +{  \mathbf{b}_i^k}\right) -\frac{1}{n}\sum_{i=1}^{n}\nabla f_i(\bm{\theta}_i^{k}) \right\|^2\right] \\
    &\leq \frac{d{\sigma}^2}{r} + \frac{\beta^2}{\gamma} + \frac{4(n-r)^2}{n^3} \sum_{i=1}^{n} \left\| \nabla f_i(\bm{\theta}_i^{k})\right\|^2.
\end{align}
\end{lemma}
\begin{lemma}
\label{lema:network_error}
Assume $k=t\tau + s$, the expected network error at iteration $k$ is bounded as follows:
\begin{align}
    \nonumber &\mathbb{E}\left[\frac{1}{r} \sum_{i\in\Omega_k}\left\|\hat{\bm{\theta}}^{k} - \bm{\theta}_i^{k}\right\|^2\right] \leq  2s^2\eta^2\Big( \frac{d{\sigma}^2(r+1)}{r} + \frac{2\beta^2}{\gamma}\Big)\\
    & \ \  + 4s\eta^2\frac{ 2(n-r)^2 + n^2}{n^3} \sum_{h=0}^{s-1}\sum_{i=1}^{n} \left\| \nabla f_i(\bm{\theta}_i^{t\tau+h})\right\|^2.
\end{align}
\end{lemma}
\begin{IEEEproof}
According to Assumption \ref{assp:smooth}, the global loss function $f$ is $L$-smooth. Let ${\mathcal{G}}^k:= ({1}/{r})\sum_{r\in\Omega_k} \left(g(\bm{\theta}_i^{k}) + \mathbf{b}_i^k\right)$, we have the expectation of the objective gap between two iterations, i.e.,
\begin{align}
    \nonumber
    & \mathbb{E}\left[f(\hat{\bm{\theta}}^{k+1}) - f(\hat{\bm{\theta}}^{k})\right]\\\nonumber
    &\leq \frac{\eta^2 L}{2} \mathbb{E}\left[\left\|{  {\mathcal{G}}^k }\right\|^2\right]  -\eta\mathbb{E}\left[\frac{1}{r}\sum_{i\in\Omega_k} \langle \nabla f(\hat{\bm{\theta}}^{k}), \mathbb{E}\left[ g(\bm{\theta}_i^{k}) + \mathbf{b}_i^k\right]\rangle\right]
    \\\nonumber
    & = \frac{\eta^2 L}{2} \mathbb{E}\left[\left\|{  {\mathcal{G}}^k }\right\|^2\right] -\eta\mathbb{E}\left[\frac{1}{r}\sum_{i\in\Omega_k} \langle \nabla f(\hat{\bm{\theta}}^{k}), \nabla f_i(\bm{\theta}_i^{k})\rangle \right]
    \\\nonumber
    & \leq \frac{\eta^2 L}{2} \mathbb{E}\left[\left\|{  {\mathcal{G}}^k }\right\|^2\right] -\frac{\eta}{2}\|\nabla f(\hat{\bm{\theta}}^{k})\|^2 - \frac{\eta}{2n}\sum_{i=1}^n \|\nabla  f_i(\bm{\theta}_i^{k})\|^2 
    \\\label{eqn:lossgap}
    &\ \ + \frac{\eta}{2}\mathbb{E} \left[\frac{1}{r}\sum_{i\in\Omega_k}\left\| \nabla f_i(\hat{\bm{\theta}}^{k})- \nabla f_i(\bm{\theta}_i^{k})\right\|^2\right],
\end{align}
where we use the inequality $-2\langle\mathbf{a}, \mathbf{b}\rangle = \|\mathbf{a}-\mathbf{b}\|^2 -\|\mathbf{a}\|^2 - \|\mathbf{b}\|^2$ for any two vectors $\mathbf{a},\mathbf{b}$. After minor rearranging, it is easy to show
\begin{align}
\label{eqn:2-4term}
    \nonumber &\mathbb{E}\left[\|\nabla f(\hat{\bm{\theta}}^{k})\|^2\right] 
    \leq \frac{2}{\eta} \mathbb{E}\left[f(\hat{\bm{\theta}}^{k}) -f(\hat{\bm{\theta}}^{k+1})\right]  + {L\eta } \mathbb{E}\left[\left\|{  {\mathcal{G}}^k }\right\|^2\right]\\
    &\ \  -\frac{1}{n}\sum_{i=1}^n \|\nabla  f_i(\bm{\theta}_i^{k})\|^2 + L^2 \mathbb{E} \left[\frac{1}{r}\sum_{i\in\Omega_k}\left\|\hat{\bm{\theta}}^{k} - \bm{\theta}_i^{k}\right\|^2\right].
\end{align}
By Lemma~\ref{lema:unbiased_update} and Lemma~\ref{lema:network_error}, we have that the last three terms of \eqref{eqn:2-4term} is bounded by $B_k$, which is
\begin{align*}
\label{B_k}
    \nonumber B_k & \geq \frac{4\eta L(n-r)^2 + (\eta L -1) n^2}{n^3} \sum_{i=1}^{n} \left\| \nabla f_i(\bm{\theta}_i^{k})\right\|^2
    \\
   \nonumber & \ \ +  \frac{ s\eta^2L^2(2(n-r)^2 + n^2)}{n^3}\sum_{h=0}^{s-1}\sum_{i=1}^{n} \left\| \nabla f_i(\bm{\theta}_i^{t\tau+h})\right\|^2
    \\
    & \ \ + 2\eta^2 L^2 s^2 \left(d\sigma^2 \left(1+\frac{1}{r}\right) + \frac{2\beta^2}{\gamma}\right) + {\eta L}\left(\frac{d\sigma^2}{r} + \frac{\beta^2}{\gamma}\right).
\end{align*}
Then, taking the total expectation and averaging of \eqref{eqn:2-4term} over all iterations, we have
\begin{equation}
\label{eqn:b_k}
   \mathbb{E}\left[\frac{1}{K}\sum_{k=0}^{K-1} \|\nabla f(\hat{\bm{\theta}}^{k})\|^2 \right] 
   \leq \frac{2(f(\bm{\theta}^{0}) -f^*)}{\eta K} + \frac{1}{K}\sum_{k=0}^{K-1}B_k,
\end{equation}
where we use the fact that $f(\bm{\theta}^{K}) \geq f^*$. Next, our goal is to find the upper bound of $({1}/{K}) \sum_{k=0}^{K-1}{B_k}$. Note that
\begin{align}
    \nonumber
    \frac{1}{K}\sum_{k=0}^{K-1}{B_k} &\leq 
    \frac{4\eta L(n-r)^2 + (\eta L -1) n^2}{Kn^3} \sum_{k=0}^{K-1}\sum_{i=1}^{n} \left\| \nabla f_i(\bm{\theta}_i^{k})\right\|^2 
    \\\nonumber
    &\ \ +  {\tau^2} \eta^2 L^2 \frac{n^2+ 2(n-r)^2}{K n^3}\sum_{k=0}^{K-1}\sum_{i=1}^{n} \left\| \nabla  f_i(\bm{\theta}_i^{k})\right\|^2 
    \\\nonumber
    &\ \ + \frac{\eta^2 L^2}{3} (\tau-1)(2\tau-1) \left(d\sigma^2 \left(1+\frac{1}{r}\right)+ \frac{2\beta^2}{\gamma}\right) \\\label{sum_bk}
    & \ \   + \frac{\eta L}{2}\left(\frac{d\sigma^2}{r} + \frac{\beta^2}{\gamma}\right)
\end{align}
based on the fact that $1^2 + \dots + n^2 = {n(n+1)(2n+1)}/{6}$. Since we have 
\begin{align*}
    &{\tau^2} \eta^2 L^2 \frac{n^2+ 2(n-r)^2}{K n^3} + \frac{4\eta L(n-r)^2 + (\eta L -1) n^2}{Kn^3} \\
    & \leq \frac{(3\tau^2\eta^2 L^2+5\eta L -1) n^2}{Kn^3},
\end{align*}
then if the learning rate $\eta$ satisfies that $ 5\eta L + 3\tau^2\eta^2 L^2\leq 1$, we can finally obtain a constant bound for $ ({1}/{K})\sum_{k=1}^{K}{B_k}$, i.e.,
\begin{align*}
    \frac{1}{K} \sum_{k=0}^{K-1}{B_k} &\leq \frac{\eta Ld\sigma^2}{2r} \left(\frac{2\eta L(r+1)}{3}(\tau-1)(2\tau-1) + 1\right)\\
    & \ \ + \frac{\eta L\beta^2}{2\gamma} \left(\frac{4\eta L}{3}(\tau-1)(2\tau-1)  +1\right)
\end{align*}
Substituting the expression of $({1}/{K})\sum_{k=0}^{K-1}B_k$ back to \eqref{eqn:b_k}, we finally obtain Theorem~\ref{thrm:non-convex}.
\end{IEEEproof}

\subsection{Proof of Lemma~\ref{thrm:convex}}\label{proof:convex}
\begin{IEEEproof}
According to Assumption~\ref{assp:convex}, \eqref{eqn:lossgap} can be written as
\begin{align}
\label{eqn:1-4term}
    \nonumber &\mathbb{E}\left[ f(\hat{\bm{\theta}}^{k+1})\right]  \leq  \eta\lambda f^* + (1-{\eta \lambda}) f(\hat{\bm{\theta}}^{k})+ \frac{\eta^2 L}{2} \mathbb{E}\left[\left\|{  {\mathcal{G}}^k }\right\|^2\right]
    \\
    &+ \frac{\eta L^2}{2} \mathbb{E} \left[\frac{1}{r}\sum_{i\in\Omega_k}\left\|\hat{\bm{\theta}}^{k} - \bm{\theta}_i^{k}\right\|^2\right]   -\frac{\eta}{2n}\sum_{i=1}^n \|\nabla  f_i(\bm{\theta}_i^{k})\|^2 ,
\end{align}
Let the last three terms in \eqref{eqn:1-4term} be bounded by $C_k$, which is equivalent to ${\eta}B_k/{2}$. Taking the total expectation and averaging over $K$ iterations based on \eqref{eqn:lossgap}, one can obtain
\begin{multline}
    \label{eqn:convex}
   \mathbb{E}\left[\frac{1}{K} \sum_{k=0}^{K-1}  f(\hat{\bm{\theta}}^{k+1})  - f^*\right]  \leq
    \frac{1-{\eta \lambda}}{K \eta \lambda}\left( f(\bm{\theta}^{0}) -  f^* \right) \\
    + \frac{1}{K{\eta\lambda}} \sum_{k=0}^{K-1}{C_k}.
\end{multline}
Next, our goal is to find the upper bound of $({1}/{K}) \sum_{k=0}^{K-1}{C_k}$. Based on \eqref{sum_bk}, we have
\begin{align*}
    \frac{1}{K}\sum_{k=0}^{K-1}{C_k} &\leq 
    \frac{4\eta^2 L(n-r)^2 + \eta(\eta L -1) n^2}{2Kn^3} \sum_{k=0}^{K-1}\sum_{i=1}^{n} \left\| \nabla f_i(\bm{\theta}_i^{k})\right\|^2 
    \\\nonumber
    &\ \ +  {\tau^2} \eta^3 L^2 \frac{n^2+ 2(n-r)^2}{2K n^3}\sum_{k=0}^{K-1}\sum_{i=1}^{n} \left\| \nabla  f_i(\bm{\theta}_i^{k})\right\|^2 
    \\\nonumber
    &\ \ + \frac{\eta^3 L^2}{6} (\tau-1)(2\tau-1) (d\sigma^2 (1+1/r)+ 2\beta^2/\gamma) \\\label{sum_bk}
    & \ \   + \frac{\eta^2 L}{4}(d\sigma^2/r + \beta^2/\gamma)
\end{align*}
If the learning rate $\eta$ satisfies that $ 5\eta L + 3\tau^2\eta^2 L^2\leq 1$, we obtain
\begin{multline*}
    \frac{1}{K} \sum_{k=0}^{K-1}{C_k} \leq \frac{\eta^2 Ld\sigma^2}{4r} \left(\frac{2\eta L(r+1)}{3}(\tau-1)(2\tau-1) + 1\right)\\
     + \frac{\eta^2 L\beta^2}{4\gamma} \left(\frac{4\eta L}{3}(\tau-1)(2\tau-1)  +1\right)
\end{multline*}
Theorem \ref{thrm:convex} follows by substituting the expression of $ ({1}/{K})\sum_{k=0}^{K-1}{C_k}$ back to \eqref{eqn:convex}.
\end{IEEEproof}

\subsection{Proof of Lemma~\ref{lema:unbiased_update}}\label{proof:unbiased_update}
\begin{IEEEproof}
To simplify the notation, we set $ \mathcal{G}^k := ({1}/{r})\sum_{i\in\Omega_k}\left(g(\bm{\theta}_i^{k}) +\mathbf{b}_i^k\right)$. Given Assumption~\ref{assp:local_unbiased}, we have
\begin{align*}
    \mathbb{E}\left[ \mathcal{G}^k\right] 
    &= \sum_{\substack{\Omega\in[n],\\ |\Omega|=r}}P_r(\Omega_k=\Omega)\left( \frac{1}{r}\sum_{i\in\Omega_k}\mathbb{E}\left[ g(\bm{\theta}_i^{k}) +{  \mathbf{b}_i^k}\right]\right)\\
    &=\frac{1}{r} \frac{1}{{n \choose r}} {n-1 \choose r-1}\sum_{i=1}^{n} \nabla f_i(\bm{\theta}_i^{k})
    = \frac{1}{n}\sum_{i=1}^{n} \nabla f_i(\bm{\theta}_i^{k}).
\end{align*}
Here, $\mathbb{E}[\mathbf{b}_i^k]=\mathbf{0}_d$ since $\mathbf{b}_i^k\sim \mathcal{N}(0, \sigma^2\mathbf{I}_d)$. let $\overline{\mathcal{G}^k}:= \mathbb{E}\left[ \mathcal{G}^k\right]$, we have
\begin{align*}
    &\mathbb{E}\left[ \left\|\mathcal{G}^k - \overline{\mathcal{G}^k}\right\|^2\right]\\
    & = \mathbb{E}\left[ \left\| \frac{1}{r}\sum_{i\in\Omega_k}g(\bm{\theta}_i^{k}) - \nabla f_i(\bm{\theta}_i^{k})\right\|^2+ \left\| \frac{1}{r}\sum_{i\in\Omega_k}\mathbf{b}_i^k \right\|^2\right] \\
    &\ \  + \mathbb{E}\left[ \left\| \frac{1}{r}\sum_{i\in\Omega_k}\nabla f_i(\bm{\theta}_i^{k}) - \frac{1}{n}\sum_{i=1}^{n}  \nabla f_i(\bm{\theta}_i^{k})\right\|^2\right]
    \\
    &\leq \frac{d\sigma^2}{r} +
    \sum_{\substack{\Omega\in[n],\\ |\Omega|=r}}P_r(\Omega_k=\Omega)
    \frac{1}{r} \sum_{i\in\Omega_k}\mathbb{E}\left[\left\| g(\bm{\theta}_i^{k}) -  \nabla f_i(\bm{\theta}_i^{k})\right\|^2\right]\\
    & \ \  + 2\sum_{\substack{\Omega\in[n],\\ |\Omega|=r}}P_r(\Omega_k=\Omega) \left( \frac{1}{r}-\frac{1}{n}\right)^2 r \sum_{i\in\Omega_k} \left\|  \nabla f_i(\bm{\theta}_i^{k}) \right\|^2\\
    & \ \ + 2\sum_{\substack{\Omega\in[n],\\ |\Omega|=r}}P_r(\Omega_k=\Omega) \frac{1}{n^2}(n-r)\sum_{i\notin\Omega_k}\left\|  \nabla f_i(\bm{\theta}_i^{k}) \right\|^2
    \\
    &\leq \frac{d\sigma^2}{r} +
    \frac{1}{n}\sum_{i=1}^{n}\mathbb{E}\left[\left\| g(\bm{\theta}_i^{k}) -  \nabla f_i(\bm{\theta}_i^{k})\right\|^2\right]\\
    & \ \  + \frac{2(n-r)}{n^2}\frac{1}{{n \choose r}}\left({n \choose r} - {n-1 \choose r-1}\right) \sum_{i=1}^{n} \left\|\nabla f_i(\bm{\theta}_i^{k})\right\|^2\\
    & \ \  + \frac{2(n-r)^2}{rn^2}\frac{1}{{n \choose r}} {n-1 \choose r-1}\sum_{i=1}^{n} \left\|\nabla f_i(\bm{\theta}_i^{k})\right\|^2
    \\
    & \leq \frac{d\sigma^2}{r} + \frac{\beta^2}{\gamma} + \frac{4(n-r)^2}{n^3} \sum_{i=1}^{n} \left\| \nabla f_i(\bm{\theta}_i^{k})\right\|^2.
\end{align*}
where we use the independence of Gaussian noise and Assumption~\ref{assp:bounded_divergence}. Note that based on Assumption~\ref{assp:bounded_divergence}, we have
\begin{align*}
    \mathbb{E}\left\| g(\bm{\theta}_i^{k}) - \nabla f_i(\bm{\theta}_i^{k})\right\|^2 &=\frac{1}{\gamma^2} \left\|\sum_{\xi_i\in X_i} \nabla f_i(\bm{\theta}_i^{k},\xi_i) - \nabla f_i(\bm{\theta}_i^{k})\right\|^2\\
    &= \frac{1}{\gamma^2} \sum_{\xi_i\in X_i}\left\| \nabla f_i(\bm{\theta}_i^{k},\xi_i) - \nabla f_i(\bm{\theta}_i^{k})\right\|^2\\
    &\leq \frac{\beta^2}{\gamma},
\end{align*}
due to the independence of random variable $\xi_i$.
\end{IEEEproof}

\subsection{Proof of Lemma~\ref{lema:network_error}}\label{proof:network_error}
\begin{IEEEproof}
Since $k=t\tau + s$ and all devices in $\Omega_k$ start from the same model received from the server $\bm{\theta}^{t\tau}$ to update, i.e., $ \hat{\bm{\theta}}^{t\tau} =\bm{\theta}_i^{t\tau}=\bm{\theta}^{t\tau}, \forall i\in\Omega_k$. For device $i\in\Omega_k$, we have
\begin{equation}
    \bm{\theta}_i^{k} = \bm{\theta}_i^{t\tau} - \eta \sum_{h=0}^{s} g(\bm{\theta}_i^{t\tau +h}) + \mathbf{b}_i^{t\tau +h}.
\end{equation}
Given that $ \hat{\bm{\theta}}^{k}=(1/r) \sum_{i\in\Omega_k} \bm{\theta}_i^{k}$, one yields $\forall j\in\Omega_k$,
\begin{align*}
    \left\|\hat{\bm{\theta}}^{k} - \bm{\theta}_j^{k}\right\|^2 & \leq {2 \eta^2} \left\|\frac{1}{r}\sum_{i\in\Omega_k}\sum_{h=0}^{s-1} g(\bm{\theta}_i^{t\tau +h}) + \mathbf{b}_i^{t\tau +h} \right\|^2 
    \\
    &\ \  + 2 \eta^2\left\|\sum_{h=0}^{s-1} g(\bm{\theta}_j^{t\tau +h}) + \mathbf{b}_j^{t\tau +h}\right\|^2\\
    & \leq 
    {2 s\eta^2} \sum_{h=0}^{s-1} \left\|\frac{1}{r}\sum_{i\in\Omega_k} g(\bm{\theta}_i^{t\tau +h}) + \mathbf{b}_i^{t\tau +h} \right\|^2 \\
    & \ \ + 2s \eta^2\sum_{h=0}^{s-1}\left\| g(\bm{\theta}_j^{t\tau +h}) + \mathbf{b}_j^{t\tau +h}\right\|^2
\end{align*}
where we use the inequality $ \|\sum_{i=1}^{n} \mathbf{a}_i\|^2 \leq n\sum_{i=1}^{n}\|\mathbf{a}_i\|^2 $. By Lemma~\ref{lema:unbiased_update} and the fact that $ \mathbb{E}[(\mathbf{X} - \mathbb{E}[\mathbf{X}])^2] =\mathbb{E}[\mathbf{X}^2] -\mathbb{E}[\mathbf{X}]^2$, we have that 
\begin{align*}
    & \mathbb{E}\left[\left\|\frac{1}{r}\sum_{i\in\Omega_k} g(\bm{\theta}_i^{t\tau +h}) + \mathbf{b}_i^{t\tau +h} \right\|^2\right] \\
    & \leq 
    \frac{d{\sigma}^2}{r}+ \frac{\beta^2}{\gamma} + \frac{4(n-r)^2+n^2}{n^3} \sum_{i=1}^{n} \left\| \nabla f_i(\bm{\theta}_i^{t\tau+h})\right\|^2,
\end{align*}
which is not related to the index of device $j$. Given that
\begin{align*}
    \left\| g(\bm{\theta}_j^{t\tau +h}) + \mathbf{b}_j^{t\tau +h}\right\|^2\leq  d\sigma^2 + \frac{\beta^2}{\gamma} + \left\| \nabla f_j(\bm{\theta}_j^{t\tau +h}) \right\|^2.
\end{align*}
Thus, the expected network error at iteration $k$ is
\begin{align*}
    & \mathbb{E}\left[ \frac{1}{r} \sum_{j\in\Omega_k}\left\|\hat{\bm{\theta}}^{k} - \bm{\theta}_j^{k}\right\|^2 \right]  \leq  2s\eta^2 \Big( \frac{sd{\sigma}^2(r+1)}{r} + \frac{2s\beta^2}{\gamma}
    \\
    & \ \  + \frac{ 4(n-r)^2 + 2n^2}{n^3} \sum_{h=0}^{s-1}\sum_{i=1}^{n} \left\| \nabla f_i(\bm{\theta}_i^{t\tau+h})\right\|^2\Big),
\end{align*}
and Lemma~\ref{lema:network_error} is finally obtained by relaxing the constant of the second term.
\end{IEEEproof}

\end{document}